\documentclass[11pt]{article}
\usepackage{amsgen,amsmath,amstext,amsbsy,amsopn,amssymb,mathabx,mathrsfs,amsthm,bm,bbm,romannum,outlines,enumitem,natbib,xr,multirow}
\usepackage{caption}
\usepackage[dvips]{graphicx}
\graphicspath{{figures/}}
\usepackage{subcaption}
\usepackage{diagbox}
\usepackage[ruled,vlined,linesnumbered,resetcount]{algorithm2e}

\usepackage{hyperref}
\hypersetup{
	colorlinks=true,
	linkcolor=red,
	urlcolor=blue,
	citecolor=blue
}
\usepackage[depth=2]{bookmark}

\usepackage[margin=1in]{geometry}
\usepackage[utf8]{inputenc}

\usepackage{booktabs}
\usepackage{longtable}
\usepackage{array}
\usepackage{multirow}
\usepackage[table]{xcolor}
\usepackage{wrapfig}
\usepackage{float}
\usepackage{colortbl}
\usepackage{pdflscape}
\usepackage{tabu}
\usepackage{threeparttable}
\usepackage{threeparttablex}
\usepackage[normalem]{ulem}
\usepackage{makecell}

\newtheorem{thm}{Theorem}
\newtheorem{lem}[thm]{Lemma}
\newtheorem{prop}[thm]{Proposition}
\newtheorem{coro}[thm]{Corollary}

\theoremstyle{definition}

\newtheorem{asm}{Assumption}

\newtheorem{rmk}{Remark}

\theoremstyle{remark}

\newcommand{\ba}{\bm{a}}

\newcommand{\bg}{\bm{g}}

\newcommand{\bx}{\bm{x}}

\newcommand{\Ib}{\mathbf{I}}

\newcommand{\Kb}{\mathbf{K}}

\newcommand{\bO}{\bm{O}}

\newcommand{\bX}{\bm{X}}

\newcommand{\bmeta}{\bm{\eta}}
\newcommand{\balpha}{\bm{\alpha}}

\newcommand{\blambda}{\bm{\lambda}}

\newcommand{\bmu}{\bm{\mu}}

\newcommand{\bbR}{\mathbb{R}}
\newcommand{\bbE}{\mathbb{E}}

\newcommand{\bbP}{\mathbb{P}}

\newcommand{\cC}{\mathcal{C}}
\newcommand{\cD}{\mathcal{D}}

\newcommand{\cG}{\mathcal{G}}
\newcommand{\cH}{\mathcal{H}}

\newcommand{\cN}{\mathcal{N}}
\newcommand{\cO}{\mathcal{O}}

\newcommand{\cV}{\mathcal{V}}

\newcommand{\cX}{\mathcal{X}}

\newcommand\smallcO{
	\mathchoice
	{{\scriptstyle\mathcal{O}}}% \displaystyle
	{{\scriptstyle\mathcal{O}}}% \textstyle
	{{\scriptscriptstyle\mathcal{O}}}% \scriptstyle
	{\scalebox{.5}{$\scriptscriptstyle\mathcal{O}$}}%\scriptscriptstyle
}

\newcommand{\bzero}{{\mathbf{0}}}	% zero vector
\newcommand{\bone}{{\mathbf{1}}}	% all-one vector
\newcommand{\bbone}{{\mathbbm{1}}}	% indicator

\newcommand{\EIF}{\mathrm{EIF}}		% efficient influence function
\newcommand{\sfvar}{\mathsf{Var}}	% variance

\newcommand{\sfsign}{\mathsf{sign}}

\newcommand\indep{\protect\mathpalette{\protect\independenT}{\perp}}
\def\independenT#1#2{\mathrel{\rlap{$#1#2$}\mkern2mu{#1#2}}}	% independence

\begin{title}
	{ Rejoinder: Learning Optimal Distributionally Robust Individualized Treatment Rules }
\end{title}
\author{
	Weibin Mo, Zhengling Qi and Yufeng Liu\thanks{Weibin Mo and Zhengling Qi are co-first authors for the paper. Weibin Mo is a Ph.D. student, Department of Statistics and Operations Research, University of North Carolina at Chapel Hill, Chapel Hill, NC 27599, USA. E-mail: \href{mailto:harrymok@email.unc.edu}{harrymok@email.unc.edu}. Zhengling Qi is Assistant Professor, Department of Decision Sciences, George Washington University, Washington, D.C. 20052, USA. E-mail: \href{mailto:qizhengling@gwu.edu}{qizhengling@gwu.edu}. Yufeng Liu is Professor, Department of Statistics and Operations Research, Department of Genetics, Department of Biostatistics, Carolina Center for Genome Science, Lineberger Comprehensive Cancer Center, University of North Carolina at Chapel Hill, NC 27599, USA. E-mail: \href{mailto:yfliu@email.unc.edu}{yfliu@email.unc.edu}.}
}
\date{}
\begin{document}
	\pagenumbering{arabic}
	\maketitle
%	\newpage
	% \pdfbookmark[<level>]{<title>}{<dest>}
%	\pdfbookmark[section]{\contentsname}{toc}
%	\tableofcontents
	
	%\newpage
	
%	\begin{abstract}
%		
%	\end{abstract}
%	\newpage
	\setcounter{page}{1}
	
	We thank the opportunity offered by editors for this discussion and the discussants for their insightful comments and thoughtful contributions. We also want to congratulate \citet{kallus2020more} for his inspiring work in improving the efficiency of policy learning by retargeting. Motivated from the discussion in \citet{dukes2020discussion}, we first point out interesting connections and distinctions between our work and \cite{kallus2020more} in Section \ref{sec:compare}. In particular, the assumptions and sources of variation for consideration in these two papers lead to different research problems with different scopes and focuses. In Section \ref{sec:evaluation}, following the discussions in \citet{li2020discussion,liang2020discussion}, we also consider the efficient policy evaluation problem when we have some data from the testing distribution available at the training stage. We show that under the assumption that the sample sizes from training and testing are growing in the same order, efficient value function estimates can deliver competitive performance. We further show some connections of these estimates with existing literature. However, when the growth of testing sample size available for training is in a slower order, efficient value function estimates may not perform well anymore. In contrast, the requirement of the testing sample size for DRITR is not as strong as that of efficient policy evaluation using the combined data. Finally, we highlight the general applicability and usefulness of DRITR in Section \ref{sec:app}.
	
	\newpage
	
	\section{Efficiency and Robustness} \label{sec:compare}
	
	The discussion in \citet{dukes2020discussion} highlighted the importance of leveraging relevant data when inferring which treatment to assign. In particular, the covariate weight functions considered in DRITR and retargeted policy learning can imply different ways of utilizing relevant data during training, and different target populations during testing. In this section, we further clarify the differences and connections between DRITR and retargeted policy learning.

	DRITR and retargeted policy learning can be distinct from each other in terms of the following two main perspectives:
	\begin{enumerate}[label=(\Roman*)]
		\item The assumptions used in these two papers are different. If the true \textit{conditional treatment effect (CTE)} function $ C(\bx) $ induces a globally optimal ITR $ \bx \mapsto \sfsign[C(\bx)] $ that belongs to the class $ \cD $, then policy learning over $ \cD $ is not sensitive to covariate reweighting/retargeting, as were discussed in \citet[Remark 1]{mo2020learning} and \citet[Lemma 2.1]{kallus2020more}. In particular, \citet{kallus2020more} referred this case as $ \cD $ being \textit{correctly specified}, and focused on this case to obtain the \textit{efficient} covariate weighting function. In contrast, \citet{mo2020learning} studied the learning problem over a restricted ITR class $ \cD $ that is \textit{misspecified} for the globally optimal ITR, and optimized the worst-case reweighted value function as a \textit{robust} objective;
		
		\item The sources of variation considered in these two papers are also different. The optimality criteria for an efficient covariate weight function in \citet{kallus2020more} focuses on reducing the \textit{conditional-on-covariate} variance of the weighted outcome, \textit{i.e.} the variance explained by $ (A,Y)|\bX $ as in their Equation (6). In contrast, \citet{mo2020learning} considered the robust criteria due to \textit{covariate} variations. In the formulation of DRITR, the effect of $ (A,Y)|\bX $ is absorbed into the CTE function $ C(\bX) = \bbE[Y(1)-Y(-1)|\bX] $ as a conditional mean function in $ \bX $. The DR-value function in \citet[Equation (4)]{mo2020learning} robustifies the underlying covariate distribution for evaluation of $ C(\bX) $.
	\end{enumerate}
	
	Two distinct assumptions on the ITR class mentioned above can explain different goals of these two papers. When assuming a correctly specified ITR class, \citet{kallus2020more} can leverage the retargeting invariance property for efficiency improvement. In contrast, when allowing the misspecified ITR class, \citet{mo2020learning} focused on the worst case of covariate changes to carry out the robust policy optimization for generalizability. The phenomenon that different model assumptions result in different goals can be remotely analogous to semiparametric inference \citep{robins2000profile}. %\QZL{Not sure if the following is a good analogy} 
	Specifically, when nuisance models are correct, the semiparametric \textit{efficient} estimate can be obtained. When either one of but not both nuisance models are correct, a doubly \textit{robust} estimate remains consistent. However, even for semiparametric inference, the goals of efficiency and robustness may not coexist for a specific estimate. For example, a generic construction of multiply robust estimate for factorized likelihood models is generally not semiparametric efficient under any model assumptions \citep{molina2017multiple}. It depends on the main focuses of applications to choose either a semiparametric efficient estimate or a multiply robust estimate. Analogously, the use of retargeted policy learning or DRITR also depends on the goal for \textit{efficiency} or \textit{robustness}, subject to the practitioners' optimistic or pessimistic beliefs on the working class of ITRs to learn from. In particular, DRITR is \textit{robust} to potential covariate changes under the misspecified ITR class assumption.
	
	The distinctions on source of variation mentioned in (\Romannum{2}) characterize two different types of research questions. When questions are related to the variations of $ (A,Y)|\bX $, such as limited overlap and heteroscedasticity, retargeting weights in \citet{kallus2020more} can provide an optimal way to control conditional-on-covariate variances \citep{crump2006moving,crump2009dealing}. However, such optimal weights cannot control the variances of covariates themselves. %, such as the variance of conditional mean corresponding to \citet[Equation (5)]{kallus2020more} and the effect of estimating weights \citep[Theorem 5.1]{crump2006moving}, cannot be controlled well. 
	Therefore, retargeted policy learning  may generally work well for the case that conditional-on-covariate variances are the estimation bottleneck, while covariate variations can be ignored. Such an example can be found in \citet[Section 4]{athey2020policy}. 
%	where the CTE function that contributes to the variance-over-covariate term should obey $ \Theta(n^{-1/2}) $ scaling. In that case, the regret lower bound can be characterized by the conditional-on-convariate variance at the $ \sqrt{n} $-scale, so that their policy learning is minimax optimal in this sense. 
	Besides ignoring variances from covariates, the violation of retargeting invariance, \textit{i.e.} misspecifying the ITR class, can also contribute biases due to reweighting. \citet[Section 5.2]{kallus2020more} also pointed out that the levels of ITR class misspecification and covariate changes need to be assumed mild when applying retargeting policy learning.
%	argued that this case should not be the main problem to handle in their approach. 
%	Putting together the concern on biases and variances due to covariate variations, the scope of research questions for retargeted policy learning should be mainly driven by variations from treatments and outcomes conditional on covariates, but NOT by covariate variations.
	
	In contrast to the scope of retargeted policy learning, DRITR has an explicit focus on covariate changes. This was motivated from the fact that the challenges of generalizing causal estimands are mainly due to covariate changes, and reweighting can correctly target the testing population of interest \citep{stuart2011use}. Without prior information on the testing population at the training stage, DRITR took the worst-case reweighting scheme to guarantee robust performance. A similar strategy was also leveraged by \citet{zhao2019robustifying}. However, given that reweighting in this case mainly aims for covariate-change correction, DRITR is not intended for handling variations from treatments and outcomes. Therefore, our formulation utilizes a nonparametric estimate of the CTE function to remove the variations from treatments and outcomes. 
	
	One potential research question is whether robust reweighting in DRITR can also handle the limited overlap and heteroscedasticity problems considered in retargeted policy learning. Unfortunately, we suspect that robust reweighting and efficient retargeting may have opposite effects. In particular, we notice that the optimal retargeting weight function from \citet{kallus2020more} is inverse-proportionate to $ \sum_{a}{\sigma^{2}(\bx,a) \over \pi_{A}(a|\bx)} $, where $ \pi_{A}(a|\bx) $ is the propensity score function. In contrast, for robust reweighting, \citet{qi2019estimation} studied a modified version of DRITR that focused on the dual formulation. Although they did not explicitly link to the robust reweigting, the resulting robust weight function in \citet{qi2019estimation} is increasing in the variance function $ \sigma^{2}(\bx,a) := \bbE(Y|\bX=\bx,A=a) $. Consequently, the robust weight function may not be compatible with the retargeting weight function. This suggests that DRITR and retargeted policy learning may need to be utilized in different scenarios. It may be interesting to study a combined version of retargeted policy learning and DRITR that can enjoy both efficiency and robustness.
	
	To conclude this section, we make a comparison between DRITR and retargeted policy learning in the following Table \ref{tab:compare}. 
	\begin{table}[!htp]
		\centering
		\begin{tabular}{l|c|c}
			\hline\hline
			& \textbf{DRITR} & \textbf{Retargeted Policy Learning}\\
			\hline\hline
			\textit{Assumption} & $ \{ \bx \mapsto \sfsign[C(\bx)] \} \nsubseteq \cD $ & $ \{ \bx \mapsto \sfsign[C(\bx)] \} \subseteq \cD $\\
			\hline
			\textit{Source of Variation} & covariate $ \bX $ & conditional-on-covariate $ (A,Y)|\bX $\\
			\hline
			\textit{Weight Dependency} & 
			\begin{tabular}{c}
				CTE function $ C(\bx) $ \\ 
				underlying ITR $ d(\bx) $
			\end{tabular} & 
			\begin{tabular}{c}
				propensity score function $ \pi_{A}(a|\bx) $ \\
				variance function $ \sigma^{2}(\bx,a) $
			\end{tabular} \\
			\hline
			\textit{Optimality} & maximizing worst-case value & minimizing conditional-on-covariate variance \\
			\hline
			\textit{Main Applications} & covariate changes & limited overlap and heteroscedasticity\\
			\hline\hline
		\end{tabular}
		\caption{Comparison of DRITR \citep{mo2020learning} and Retargeted Policy Learning \citep{kallus2020more}}
		\label{tab:compare}
	\end{table}
	
	\section{Efficient Policy Evaluation under  Specific Covariate Changes} \label{sec:evaluation}
	DRITR aims for performance guarantee in presence of general covariate changes. It assumes no access to any information from the testing distribution at the training stage. When a small set of calibrating data is available from the testing distribution, such information is only used for choosing a DR-constant to determine the final DRITR. However, the problem of combining the training and calibrating data during training as discussed in \citet{li2020discussion,liang2020discussion} is also worthwhile to study. In this section, we focus on efficient policy evaluation with training and calibrating data from a specific testing distribution. It should be highlighted that the true covariate density ratio of testing with respect to training is not readily available, and can only be inferred from the observed training and calibrating data.
	
	Consider two possible types of pooled datasets $ \cO^{(1)} = \{ \bO_{i} = (\bX_{i},A_{i},Y_{i},S_{i}) \}_{i=1}^{n} $ and $ \cO^{(2)} = \{ \bO_{i} = (\bX_{i},S_{i}A_{i},S_{i}Y_{i},S_{i}) \}_{i=1}^{n} $, where $ S_{i} = 1 $ indicates that $ \bO_{i}|(S_{i}=1) \sim \bbP_{\rm train} $, and the $ i $-th data point belongs to the training data; $ S_{i} = 0 $ indicates that $ \bO_{i}|(S_{i}=0) \sim \bbP_{\rm test} $, and the $ i $-th data point belongs to the calibrating data. For the Type-1 dataset $ \cO^{(1)} $, we observe covariates, treatment assignments and outcomes in both training and calibrating data. For the Type-2 dataset $ \cO^{(2)} $, treatment assignments and outcomes $ \{(A_{i},Y_{i}):S_{i} = 0\} $ in calibrating data are missing. Let $ Y_{i}(1) $ and $ Y_{i}(-1) $ be the potential outcomes, and denote $ Y_{i}(d) := \sum_{a \in \{1,-1\}}Y_{i}(a)\bbone[d(\bX_{i})=a] $. For a fixed ITR $ d: \cX \to \{1,-1\} $, the goal for policy evaluation is to estimate the following values of $ d $ under the specific testing distribution:
	\begin{align}
		\theta := \cV_{\rm test}(d) = \bbE_{\rm test}[Y_{i}(d)]; \quad \theta_{1} := \cV_{1,\rm test}(d) = \bbE_{\rm test}[Y_{i}(d) - Y_{i}(-d)].\label{eq:param}
	\end{align}
	In order to identify the potential outcomes from observed outcomes, we make the following assumptions \citep{rubin1974estimating}.
	
	\begin{asm}[Consistency]\label{asm:consistency}
		$ Y_{i} = Y_{i}(A_{i}) = \sum_{a \in \{1,-1\}}Y_{i}(a)\bbone(A_{i}=a) $.
	\end{asm}
	
	\begin{asm}[Exchangeability over Treatment]\label{asm:mean_a}
		For $ \bx \in \cX $, $ s \in \{1,0\} $ and $ a \in \{1,-1\} $, we have $ Y_{i}(a)\indep A_{i}|(\bX_{i} = \bx, S_{i}=s) $.
	\end{asm}

	\begin{asm}[Strong Ignorability of Treatment]\label{asm:pos_a}
		For $ \bx \in \cX $, $ s \in \{1,0\} $ and $ a \in \{1,-1\} $, we have $ \pi_{A}(\ba|\bx,s) := \bbP(A_{i}=a|\bX_{i}=\bx,S_{i}=s) \ge \tau > 0 $.
	\end{asm}

	Notice that the policy evaluation problem can be cast as a parameter estimation problem. The key challenge here is that the estimand $ \theta $ or $ \theta_{1} $ is evaluated under the target population $ \bbP_{\rm test} $ that can be different from the distribution of the observed data $ \cO^{(1)} $ or $ \cO^{(2)} $. In order to identify $ \theta $ and $ \theta_{1} $ from the pooled data $ \cO^{(1)} $ or $ \cO^{(2)} $, we consider the following \textit{mean exchangeability assumption} used in \citet{pearl2014external} to transport information from $ \bbP_{\rm train} $ to $ \bbP_{\rm test} $. 
	\begin{asm}[Mean Exchangeability over Selection] \label{asm:mean_s}
		For $ \bx \in \cX $, $ a \in \{1,-1\} $ and $ s \in \{1,0\} $, we have $ \bbE[Y_{i}(a)|\bX_{i}=\bx,S_{i}=s] = \bbE[Y_{i}(a)|\bX_{i}=\bx] := Q(\bx,a) $. 
	\end{asm}
	\noindent Assumption \ref{asm:mean_s} implies that $ \bbE[Y_{i}(1) - Y_{i}(-1)|\bX_{i},S_{i}] = \bbE[Y_{i}(1) - Y_{i}(-1)|\bX_{i}] = Q(\bX_{i},1) - Q(\bX_{i},-1) := C(\bX_{i}) $, which was the required condition in \citet[Remark 2]{mo2020learning}. Covariate changes in \citet[Assumption 1]{mo2020learning} is sufficient for Assumption \ref{asm:mean_s}. In order to identify $ \theta $ and $ \theta_{1} $ from the training and pooled data, we further consider the \textit{strong ignorability assumption} on the conditional-on-covariate selection probability function. 
	\begin{asm}[Strong Ignorability of Selection]\label{asm:pos_s}
		For $ \bx \in \cX $ and $ s \in \{1,0\} $, we have $ \pi_{S}(s|\bx) := \bbP(S_{i}=s|\bX_{i}=\bx) \ge \delta > 0 $.
	\end{asm}
	\noindent Assumption \ref{asm:pos_s} also implies the positivity of the marginal selection probability: $ \rho_{S} := \bbP(S_{i} = 1) \ge \delta $, and $ 1-\rho_{S} = \bbP(S_{i}=0) \ge \delta $. Under Assumption \ref{asm:pos_s}, we can express $ \bbE_{\rm test}(\cdot) = \bbE(\cdot|S=0) $ under $ \bbE_{\rm train}(\cdot) = \bbE(\cdot|S=1) $ by reweighting, which is given in the following Lemma \ref{lem:weight}.
	
	\begin{lem}\label{lem:weight}
		Consider the data $ (\bX,S) \in \cX \times \{1,0\} $ satisfying $ \bbP(S=s|\bX) \ge \delta > 0 $ for $ s \in \{1,0\} $. Then for any $ g:\cX \to \bbR $ such that $ \bbE[|g(\bX)|] < +\infty $, we have
		\[ \bbE[g(\bX)|S=0] = \bbE[w(\bX)g(\bX)|S=1], \]
		where
		\[ w(\bX) = {\bbP(S=1)\bbP(S=0|\bX) \over \bbP(S=0)\bbP(S=1|\bX)}. \]
		In particular, $ w(\bX) \ge 0 $ and $ \bbE[w(\bX)|S=1] = 1 $.
	\end{lem}

	\noindent Lemma \ref{lem:weight} can be regarded as a parallel version of the weighted representation in \citet[Assumption 1]{mo2020learning}, and further suggests the form of the theoretical weighting function $ w(\bx) = {\rho_{S} \over 1 - \rho_{S}}{\pi_{S}(0|\bx) \over \pi_{S}(1|\bx)} $ for $ \bx \in \cX $. 
	
	\subsection{Semiparametric Inference}\label{sec:inference}
	In order to consider estimates for $ \theta $ and $ \theta_{1} $, we first study the semiparametric inference properties for $ \theta $ and $ \theta_{1} $ based on two types of data $ \cO^{(1)} $ and $ \cO^{(2)} $ respectively. In Proposition \ref{prop:ID}, we establish the identification of $ \theta $ and $ \theta_{1} $ from the observed data. Then we derive the efficient influence functions for $ \theta $ and $ \theta_{1} $ in Theorem \ref{thm:EIF}, which is consistent with results from \citet{rudolph2017robust}, \citet{dahabreh2019efficient} and \citet{uehara2020off}. For $ \bx \in \cX $, we denote $ Q(\bx,d) := \sum_{a \in \{1,-1\}}Q(\bx,a)\bbone[d(\bx) = a] $.
	
	\begin{prop} \label{prop:ID}
		Consider the observed data $ \cO^{(1)} $ and $ \cO^{(2)} $, and the parameters $ \theta $ and $ \theta_{1} $ in (\ref{eq:param}). Under Assumptions \ref{asm:consistency}-\ref{asm:pos_s}, we have:
		\begin{enumerate}[label=(\Roman*)]
			\item 
			\begin{align*}
				\theta &= \bbE\left\{ {\bbone(S_{i}=0) \over 1- \rho_{S}}Q(\bX_{i},d) \right\}\\
				&= \bbE\Big\{ [w(\bX_{i})\bbone(S_{i}=1) + \bbone(S_{i}=0)]Q(\bX_{i},d) \Big\} && (\text{on $ \cO^{(1)} $}) \\
				&= \bbE\left\{ {w(\bX_{i})\bbone(S_{i}=1) \over \rho_{S}}Q(\bX_{i},d) \right\} && (\text{on $ \cO^{(2)} $}).
			\end{align*}
			The expressions for $ \theta_{1} $ can be obtained by replacing $ Q(\bX_{i},d) $ by $ C(\bX_{i})d(\bX_{i}) $;
		
			\item For $ \bx \in \cX $, 
			\begin{align*}
				Q(\bX_{i},d) &= \bbE\left\{ \left( {\bbone(S_{i}=1) \over \pi_{A}(A_{i}|\bX_{i},1)} + {\bbone(S_{i}=0) \over \pi_{A}(A_{i}|\bX_{i},0)} \right)\bbone[d(\bX_{i})=A_{i}]Y_{i}\middle|\bX_{i} \right\} && (\text{on $ \cO^{(1)} $}) \\
				&= \bbE\left\{ {1 \over \pi_{A}(A_{i}|\bX_{i},1)}\bbone[d(\bX_{i})=A_{i}]Y_{i}\middle|\bX_{i},S_{i}=1 \right\} && (\text{on $ \cO^{(2)} $}).
			\end{align*}
			The expressions for $ \theta_{1} $ can be obtained by replacing $ \bbone[d(\bX_{i})=A_{i}] $ by $ d(\bX_{i})A_{i} $.
		\end{enumerate}
	\end{prop}
	
	\begin{thm}[Efficient Influence Functions] \label{thm:EIF}
		Consider the observed data $ \cO^{(1)} $ and $ \cO^{(2)} $, and the parameters $ \theta $ and $ \theta_{1} $ in (\ref{eq:param}). Under Assumptions \ref{asm:consistency}-\ref{asm:pos_s} and some regularity conditions, the corresponding semiparametric \textit{efficient influence functions (EIFs)} are:
		\begin{align*}
			\begin{array}{rclclclcl}
				\EIF^{(1)}(\theta) &=& \left( {w(\bX)\bbone(S=1) \over \pi_{A}(A|\bX,1)} + {\bbone(S=0) \over \pi_{A}(A|\bX,0)} \right) & \times & \bbone[d(\bX)=A][Y - Q(\bX,A)] & + & {\bbone(S=0) \over 1-\rho_{S}}[Q(\bX,d) - \theta];\\
				\EIF^{(2)}(\theta) &=& {w(\bX)\bbone(S=1) \over \rho_{S}\pi_{A}(A|\bX,1)} & \times & \bbone[d(\bX)=A][Y - Q(\bX,A)] & + & {\bbone(S=0) \over 1-\rho_{S}}[Q(\bX,d) - \theta];\\
				\EIF^{(1)}(\theta_{1}) &=& \left( {w(\bX)\bbone(S=1) \over \pi_{A}(A|\bX,1)} + {\bbone(S=0) \over \pi_{A}(A|\bX,0)} \right) & \times & d(\bX)A[Y - Q(\bX,A)] & + & {\bbone(S=0) \over 1-\rho_{S}}[C(\bX)d(\bX) - \theta_{1}];\\
				\EIF^{(2)}(\theta_{1}) &=& {w(\bX)\bbone(S=1) \over \rho_{S}\pi_{A}(A|\bX,1)} &\times& d(\bX)A[Y - Q(\bX,A)] & + & {\bbone(S=0) \over 1-\rho_{S}}[C(\bX)d(\bX) - \theta_{1}].
			\end{array}
		\end{align*}
		Here, $ \EIF^{(k)} $ represents the EIF based on $ \cO^{(k)} $ for $ k = 1,2 $.
	\end{thm}

	\noindent Theorem \ref{thm:EIF} can be connected to existing literature. \citet[Section 4]{rudolph2017robust} obtained the EIF for $ \theta_{1} $ on the Type-2 data $ \cO^{(2)} $. \citet[Appendix D]{dahabreh2019efficient} first obtained the EIF for $ \theta $ based on the Type-2 data $ \cO^{(2)} $. Then they further argued that if Assumption \ref{asm:mean_a} is relaxed to that $ Y_{i}(a) \indep A_{i} | (\bX_{i} = \bx,S_{i} = 1) $ holds on the training data only, then the observed treatment assignments and outcomes $ \{ (A_{i},Y_{i}): S_{i} = 0 \} $ from the calibrating data cannot be used. In that case, the EIFs based on $ \cO^{(1)} $ and $ \cO^{(2)} $ are the same. \citet[Theorem 11]{uehara2020off} obtained the EIF for $ \theta $ on the Type-2 data $ \cO^{(2)} $ using a stratified sampling formulation for $ \bO_{i}|(S_{i}=1) $ and $ \bO_{i}|(S_{i}=0) $, and treating $ w(\bX_{i}) $ as a general density ratio function of covariates. Finally, we would like to point out that \citet[Lemma 5.1]{kallus2020more} considered the EIF for a fixed weight function, while the weight function $ w(\bx) $ as in Lemma \ref{lem:weight} is model-endogeneous in the sense that it corresponds to the conditional-on-covariate selection odds or the covariate density ratio in the semiparametric model. The EIF in \citet[Lemma 5.1]{kallus2020more} assumes that $ w(\bx) $ is known or the associated nuisance function $ \pi_{S}(s|\bx,a) $ is known, while the EIFs from our Theorem \ref{thm:EIF} account for additional variances contributed by estimating $ w(\bx) $.
	
	Suppose $ w(\bx) $, $ \pi_{A}(a|\bx,s) $ and $ Q(\bx,a) $ are known as oracle. Recall that $ C(\bx) = Q(\bx,1) - Q(\bx,-1) $. Denote $ n_{s} := \#\{i:S_{i}=s\} $ for $ s \in \{1,0\} $. Then we have $ n = n_{1} + n_{0} $ and $ n_{1}/n \to \rho_{S} $ as $ n \to \infty $. Theorem \ref{thm:EIF} can imply the following oracle semiparametric efficient estimates of $ \theta $ and $ \theta_{1} $ based on $ \cO^{(1)} $ and $ \cO^{(2)} $ respectively:
	\begin{align}
		\widehat{\theta}_{\rm eff}^{(1)} = \quad & {1 \over n_{1}}\sum_{i:S_{i}=1} {n_{1} \over n}w(\bX_{i}){\bbone[d(\bX_{i})=A_{i}] \over \pi_{A}(A_{i}|\bX_{i},1)}[Y_{i} - Q(\bX_{i},A_{i})] && (:= \phi_{n_{1}}^{(1)}) \nonumber\\
		+& {1 \over n_{0}} \sum_{i:S_{i} = 0}\left\{ {n_{0} \over n}{\bbone[d(\bX_{i})=A_{i}] \over \pi_{A}(A_{i}|\bX_{i},0)}[Y_{i} - Q(\bX_{i},A_{i})] + Q(\bX_{i},d) \right\} && (:= \psi_{n_{0}}^{(1)} + \theta); \nonumber\\
		\widehat{\theta}_{\rm eff}^{(2)} = \quad & {1 \over n_{1}}\sum_{i:S_{i}=1}w(\bX_{i}){\bbone[d(\bX_{i})=A_{i}] \over \pi_{A}(A|\bX,1)} [Y_{i} - Q(\bX_{i},A_{i})] && (:= \phi_{n_{1}}^{(2)}) \nonumber \\
		+& {1 \over n_{0}}\sum_{i: S_{i}=0}Q(\bX_{i},d) && (:= \psi_{n_{0}}^{(2)} + \theta); \nonumber\\
		\widehat{\theta}_{1,\rm eff}^{(1)} = \quad & {1 \over n_{1}}\sum_{i:S_{i}=1}{n_{1} \over n} w(\bX_{i}){d(\bX_{i})A_{i} \over \pi_{A}(A_{i}|\bX_{i},1)}[Y_{i} - Q(\bX_{i},A_{i})] && (:= \phi_{1,n_{1}}^{(1)}) \nonumber \\
		+& {1 \over n_{0}}\sum_{i: S_{i}=0}\left\{ {n_{0} \over n}{\bbone[d(\bX_{i})=A_{i}] \over \pi_{A}(A_{i}|\bX_{i},0)}[Y_{i} - Q(\bX_{i},A_{i})] + C(\bX_{i})d(\bX_{i}) \right\} && (:= \psi_{1,n_{0}}^{(1)} + \theta_{1}); \nonumber\\
		\widehat{\theta}_{1,\rm eff}^{(2)} = \quad & {1 \over n_{1}}\sum_{i:S_{i}=1}w(\bX_{i}){d(\bX_{i})A_{i} \over \pi_{A}(A_{i}|\bX_{i},1)} [Y_{i} - Q(\bX_{i},A_{i})] && (:= \phi_{1,n_{1}}^{(2)}) \nonumber \\
		+& {1 \over n_{0}}\sum_{i: S_{i}=0}C(\bX_{i})d(\bX_{i}) && (:= \psi_{1,n_{0}}^{(2)} + \theta_{1}). \label{eq:oracle}
	\end{align}
	Here, we use $ n_{1}/n $ and $ n_{0}/n $ to replace $ \rho_{S} $ and $ 1 - \rho_{S} $ respectively. The asymptotic variances of the oracle semiparametric estimates in (\ref{eq:oracle}) are the smallest among all \textit{regular and asymptotic linear (RAL)} estimates \citep{tsiatis2007semiparametric}, which we establish in the following Theorem \ref{thm:eff}. For $ \bx \in \cX $, $ s \in \{1,0\} $ and $ a \in \{1,-1\} $, we denote $ \sigma^{2}(\bx,s,a) := \sfvar[Y_{i}(a)|\bX_{i}=\bx,S_{i}=s] $, $ \sigma^{2}(\bx,s,d) := \sum_{a \in \{1,-1\}}\sigma^{2}(\bx,s,a)\bbone[d(\bx_{i})=a] $ and $ \pi_{A}(d|\bx,s) := \sum_{a \in \{1,-1\}}\pi_{A}(a|\bx,s) $. In order to obtain asymptotic results for all these estimates, we make the following integrability assumption.
	\begin{asm}[Squared Integrability]\label{asm:intble}
		Assume that 
		\[ \bbE[Q(\bX_{i},1)^{2}], ~ \bbE[Q(\bX_{i},-1)^{2}] < +\infty; \quad \sup_{\bx \in \cX, s \in \{1,0\}, a \in \{1,-1\}}\sigma^{2}(\bx,s,a) < +\infty. \]
	\end{asm}

	\begin{thm}[Semiparametric Efficiency]\label{thm:eff}
		Consider the observed data $ \cO^{(1)} $ and $ \cO^{(2)} $, the parameters $ \theta $ and $ \theta_{1} $ in (\ref{eq:param}), and the corresponding oracle efficient estimates in (\ref{eq:oracle}). Under Assumptions \ref{asm:consistency}-\ref{asm:intble}, we have
		\begin{align*}
			\begin{array}{lllll}
				\sqrt{n_{1}}\phi_{n_{1}}^{(1)} \stackrel{n_{1} \to \infty}{\Longrightarrow} Z_{1}^{(1)}; & \sqrt{n_{0}}\psi_{n_{0}}^{(1)} \stackrel{n_{0} \to \infty}{\Longrightarrow} Z_{0}^{(1)}; & Z_{1}^{(1)} \sim \cN(0, \nu_{\rm eff}^{(1)}); & Z_{0}^{(1)} \sim \cN(0, \zeta_{\rm eff}^{(1)}); & Z_{1}^{(1)} \indep Z_{0}^{(1)};\\
				\sqrt{n_{1}}\phi_{n_{1}}^{(2)} \stackrel{n_{1} \to \infty}{\Longrightarrow} Z_{1}^{(2)}; & \sqrt{n_{0}}\psi_{n_{0}}^{(2)} \stackrel{n_{0} \to \infty}{\Longrightarrow} Z_{0}^{(2)}; & Z_{1}^{(2)} \sim \cN(0, \nu_{\rm eff}^{(2)}); & Z_{0}^{(2)} \sim \cN(0, \zeta_{\rm eff}^{(2)}); & Z_{1}^{(2)} \indep Z_{0}^{(2)};\\
				\sqrt{n_{1}}\phi_{1,n_{1}}^{(1)} \stackrel{n_{1} \to \infty}{\Longrightarrow} Z_{11}^{(1)}; & \sqrt{n_{0}}\psi_{1,n_{0}}^{(1)} \stackrel{n_{0} \to \infty}{\Longrightarrow} Z_{10}^{(1)}; & Z_{11}^{(1)} \sim \cN(0, \nu_{1,\rm eff}^{(1)}); & Z_{10}^{(1)} \sim \cN(0, \zeta_{1,\rm eff}^{(1)}); & Z_{11}^{(1)} \indep Z_{10}^{(1)};\\
				\sqrt{n_{1}}\phi_{1,n_{1}}^{(2)} \stackrel{n_{1} \to \infty}{\Longrightarrow} Z_{11}^{(2)}; & \sqrt{n_{0}}\psi_{1,n_{0}}^{(2)} \stackrel{n_{0} \to \infty}{\Longrightarrow} Z_{10}^{(2)}; & Z_{11}^{(2)} \sim \cN(0, \nu_{1,\rm eff}^{(2)}); & Z_{10}^{(2)} \sim \cN(0, \zeta_{1,\rm eff}^{(2)}); & Z_{11}^{(2)} \indep Z_{10}^{(2)},
			\end{array}
		\end{align*}
		where
		\begin{align*}
			\begin{array}{rlrl}
				\nu_{\rm eff}^{(1)} &= \rho_{S}^{2}\bbE\left\{ w(\bX_{i})^{2}{\sigma^{2}(\bX_{i},1,d) \over \pi_{A}(d|\bX_{i},1)}\middle| S_{i} = 1\right\}; \\
				\zeta_{\rm eff}^{(1)} &= (1-\rho_{S})^{2}\bbE\left\{ {\sigma^{2}(\bX_{i},0,d) \over \pi_{A}(d|\bX_{i},0)}\middle| S_{i} = 0\right\} + \sfvar[Q(\bX_{i},d)|S_{i} = 0];\\
				\nu_{\rm eff}^{(2)} &= \bbE\left\{ w(\bX_{i})^{2}{\sigma^{2}(\bX_{i},1,d) \over \pi_{A}(d|\bX_{i},1)}\middle| S_{i} = 1\right\}; \\
				\zeta_{\rm eff}^{(2)} &= \sfvar[Q(\bX_{i},d)|S_{i} = 0];\\
				\nu_{1,\rm eff}^{(1)} &= \rho_{S}^{2}\bbE\left\{ w(\bX_{i})^{2}\sum_{a \in \{1,-1\}}{\sigma^{2}(\bX_{i},1,a) \over \pi_{A}(a|\bX_{i},1)}\middle| S_{i} = 1\right\}; \\
				\zeta_{1,\rm eff}^{(1)} &= (1-\rho_{S})^{2}\bbE\left\{ \sum_{a \in \{1,-1\}}{\sigma^{2}(\bX_{i},0,a) \over \pi_{A}(a|\bX_{i},0)}\middle| S_{i} = 0\right\} + \sfvar[C(\bX_{i})d(\bX_{i})|S_{i} = 0];\\
				\nu_{1,\rm eff}^{(2)} &= \bbE\left\{ w(\bX_{i})^{2}\sum_{a \in \{1,-1\}}{\sigma^{2}(\bX_{i},1,a) \over \pi_{A}(a|\bX_{i},1)}\middle| S_{i} = 1\right\}; \\
				\zeta_{1,\rm eff}^{(2)} &= \sfvar[C(\bX_{i})d(\bX_{i})|S_{i} = 0].
			\end{array}
		\end{align*}
		Moreover, for $ \{\alpha_{n}\} $ such that $ \alpha_{n}/\sqrt{n_{1}} \to \gamma_{1} $ and $ \alpha_{n}/\sqrt{n_{0}} \to \gamma_{0} $ as $ n \to \infty $, we have
		\begin{align*}
			\begin{array}{rll}
				\alpha_{n}(\widehat{\theta}_{\rm eff}^{(1)} - \theta) &= (\gamma_{1}+\smallcO(1)) \times \sqrt{n_{1}}\phi_{n_{1}}^{(1)} + (\gamma_{0}+\smallcO(1)) \times \sqrt{n_{0}}\psi_{n_{0}}^{(1)} &\stackrel{n_{1},n_{0} \to \infty}{\Longrightarrow} \gamma_{1}Z_{1}^{(1)} + \gamma_{0}Z_{0}^{(1)};\\
				\alpha_{n}(\widehat{\theta}_{\rm eff}^{(2)} - \theta) &= (\gamma_{1}+\smallcO(1)) \times \sqrt{n_{1}}\phi_{n_{1}}^{(2)} + (\gamma_{0}+\smallcO(1)) \times \sqrt{n_{0}}\psi_{n_{0}}^{(2)} &\stackrel{n_{1},n_{0} \to \infty}{\Longrightarrow} \gamma_{1}Z_{1}^{(2)} + \gamma_{0}Z_{0}^{(2)};\\
				\alpha_{n}(\widehat{\theta}_{1,\rm eff}^{(1)} - \theta_{1}) &= (\gamma_{1}+\smallcO(1)) \times \sqrt{n_{1}}\phi_{1,n_{1}}^{(1)} + (\gamma_{0}+\smallcO(1)) \times \sqrt{n_{0}}\psi_{1,n_{0}}^{(1)} &\stackrel{n_{1},n_{0} \to \infty}{\Longrightarrow} \gamma_{1}Z_{11}^{(1)} + \gamma_{0}Z_{10}^{(1)};\\
				\alpha_{n}(\widehat{\theta}_{1,\rm eff}^{(1)} - \theta_{1}) &= (\gamma_{1}+\smallcO(1)) \times \sqrt{n_{1}}\phi_{1,n_{1}}^{(2)} + (\gamma_{0}+\smallcO(1)) \times \sqrt{n_{0}}\psi_{1,n_{0}}^{(2)} &\stackrel{n_{1},n_{0} \to \infty}{\Longrightarrow} \gamma_{1}Z_{11}^{(2)} + \gamma_{0}Z_{10}^{(2)}.
			\end{array}
		\end{align*}
		In particular, for $ \alpha_{n} = \sqrt{n} $ (\textit{resp.} $ \sqrt{n_{1}} $ or $ \sqrt{n_{0}} $), the corresponding $ \sqrt{n} $ (\textit{resp.} $\sqrt{n_{1}}$ or $ \sqrt{n_{0}} $) asymptotic variances achieve the semiparametric $ \sqrt{n} $-(\textit{resp.} $ \sqrt{n_{1}} $-or $ \sqrt{n_{0}} $-)variance lower bounds for $ \theta $ and $ \theta_{1} $ on $ \cO^{(1)} $ and $ \cO^{(2)} $ respectively.
	\end{thm}

	We notice that the efficient estimates (\ref{eq:oracle}) can depend on the unknown nuisance functions $ w(\bx) $, $ \pi_{A}(a|\bx,s) $ and $ Q(\bx,a) $. Implementable efficient estimates of $ \theta $ and $ \theta_{1} $ are the plug-in versions with the corresponding sample-dependent nuisance function estimates. \citet{uehara2020off} took the following cross-fitting strategy \citep{chernozhukov2018double} when plugging in the nuisance function estimates: the pooled data are stratified into $ \{ i:S_{i}=1 \} $ and $ \{ i:S_{i} = 0 \} $, and the sample points within each stratum are randomly divided into $ K $ bags. For $ k \in \{1,2,\cdots,K\} $, we obtain out-of-bag estimates of the nuisance functions from the pooled dataset that rules out the $ k $-th bag of data points. When constructing the cross-fitting estimates (\ref{eq:oracle}) of $ \theta $ and $ \theta_{1} $, if the $ i $-th data point belongs to the $ k $-th bag, then it utilizes the out-of-$ k $-th-bag nuisance function estimates. Using the cross-fitting strategy, we can follow \citet[Theorem 2]{uehara2020off} and establish $ \sqrt{n} $-equivalences for plug-in efficient estimates. In the following Theorem \ref{thm:equiv}, we only consider the cross-fitting estimates for $ \theta $, and the same argument can be applied to $ \theta_{1} $.
	
	\begin{thm}[$ \sqrt{n} $-Equivalence]\label{thm:equiv}
		Consider the observed data $ \cO^{(1)} $ and $ \cO^{(2)} $ and the parameter $ \theta $ in (\ref{eq:param}). Denote $ \Phi^{(1)}(\eta) := \widehat{\theta}_{\rm eff}^{(1)} - \theta $ and $ \Phi^{(2)}(\eta) = \widehat{\theta}_{\rm eff}^{(2)} - \theta $ from (\ref{eq:oracle}) with $ \eta =  \big( w(\bx), \pi_{A}(a|\bx,s), Q(\bx,a) \big) $ as the nuisance functions. For $ k \in \{1,2,\cdots,K\} $, let $ \widehat{\eta}_{\rm cross} := \{(\widehat{w}^{(k)},\widehat{\pi}_{A}^{(k)},\widehat{Q}^{(k)})\}_{k=1}^{K} $ be the out-of-bag nuisance function estimates, and $ \Phi^{(1)}(\widehat{\eta}_{\rm cross}) $ and $ \Phi^{(2)}(\widehat{\eta}_{\rm cross}) $ be the corresponding cross-fitting versions. Define 
		\begin{align*}
			\alpha_{n}^{(2)} &:= \max_{a \in \{1,-1\}}\max_{1 \le k \le K}\left\|{\widehat{w}^{(k)}(\cdot) \over \widehat{\pi}_{A}^{(k)}(a|\cdot,1)} - {w(\cdot) \over \pi_{A}(a|\cdot,1)}\right\|_{L^{2}(\bbP)}; \\
			\alpha_{n}^{(1)} &:= \alpha_{n}^{(2)} + \max_{1 \le k \le K}\left\| \widehat{\pi}_{A}^{(k)}(1|\cdot,0) - \pi_{A}(1|\cdot,0) \right\|_{L^{2}(\bbP)}; \\
			\beta_{n} &:= \max_{a \in \{1,-1\}}\max_{1 \le k \le K}\left\|\widehat{Q}^{(k)}(\cdot,a) - Q(\cdot,a)\right\|_{L^{2}(\bbP)},
		\end{align*}
		where $ \|\cdot\|_{L^{2}(\bbP)} $ is the $ L^{2}(\bbP) $-norm with respect to the covariate vector $ \bX $. Then under Assumptions \ref{asm:consistency}-\ref{asm:intble} and that $ \alpha_{n}^{(1)},\alpha_{n}^{(2)},\beta_{n} = \smallcO_{\bbP}(1) $, we have
		\[ \sqrt{n}[\Phi^{(1)}(\widehat{\eta}_{\rm cross}) - \Phi^{(1)}(\eta)] = \cO_{\bbP}(\sqrt{n}\alpha_{n}^{(1)}\beta_{n}); \quad \sqrt{n}[\Phi^{(2)}(\widehat{\eta}_{\rm cross}) - \Phi^{(2)}(\eta)] = \cO_{\bbP}(\sqrt{n}\alpha_{n}^{(2)}\beta_{n}). \]
	\end{thm}

	\begin{rmk}
		Notice that in Theorem \ref{thm:equiv}, by the fact that $ \pi_{A}(a|\bx,s) \ge \tau $, we further have
		\[ \alpha_{n}^{(2)} \le \text{constant} \times \max_{1 \le k \le K}\left\| \widehat{w}^{(k)} - w \right\|_{L^{2}(\bbP)} + \text{constant} \times \max_{1 \le k \le K}\left\| \widehat{\pi}_{A}^{(k)}(1|\cdot,1) - \pi_{A}(1|\cdot,1) \right\|_{L^{2}(\bbP)}. \]
		Then a dominating rate of $ \alpha_{n}^{(2)} $ can be chosen as the slower one of $ \widehat{w}(\cdot) $ and $ \widehat{\pi}_{A}(1|\cdot,1) $.
	\end{rmk}

	One important implication of the remainder terms in Theorem \ref{thm:equiv} is that $ \alpha_{n}^{(1)} $, $ \alpha_{n}^{(2)} $ and $ \beta_{n} $ can be in slower orders than the usual requirement $ \smallcO_{\bbP}(n^{-1/2}) $ for negligibility. For example, if $ \alpha_{n}^{(1)} = \smallcO_{\bbP}(n^{-1/4}) $ and $ \beta_{n} = \smallcO_{\bbP}(n^{-1/4}) $, then the plug-in effect for $ \Phi^{(1)}(\widehat{\eta}_{\rm cross}) $ is negligible.

	Finally, we notice that the $ \sqrt{n} $-asymptotic results in Theorem \ref{thm:eff} rely on the implication from Assumption \ref{asm:pos_s} that $ n_{1}/n \to \rho_{S} \ge \delta $ and $ n_{0}/n \to 1-\rho_{S} \ge \delta $ as $ n \to \infty $, so that both $ n_{1} $ and $ n_{0} $ grow linearly in $ n $. When the calibrating sample size $ n_{0} $ is of a smaller order in the training sample size $ n_{1} $, the leading order terms for the estimates in (\ref{eq:oracle}) are of order $ \sqrt{n_{0}} $. We establish the corresponding asymptotic results in the following Corollary \ref{coro:eff} that are different from those in Theorem \ref{thm:eff}.

	\begin{coro}[Semiparametric Efficiency with Small Calibrating Data]\label{coro:eff}
		Consider the observed data $ \cO^{(1)} $ and $ \cO^{(2)} $, the parameters $ \theta $ and $ \theta_{1} $ in (\ref{eq:param}), and the corresponding oracle efficient estimates in (\ref{eq:oracle}). Suppose $ n_{0}/n_{1} \to 0 $ as $ n_{1},n_{0} \to \infty $. Then under Assumptions \ref{asm:consistency}-\ref{asm:mean_s}, \ref{asm:intble} and that $ \sup_{\bx \in \cX}w(\bx) < +\infty $, we have
		\begin{align*}
			\begin{array}{c}
				\sqrt{n_{0}}(\widehat{\theta}_{\rm eff}^{(2)} - \theta) \underset{n_{0}\to\infty}{\stackrel{\cD}{\longrightarrow}}\cN(0,\sfvar[Q(\bX_{i},d)|S_{i}=0]); \quad  \sqrt{n_{0}}(\widehat{\theta}_{\rm eff}^{(2)} - \theta) \underset{n_{0}\to\infty}{\stackrel{\cD}{\longrightarrow}}\cN(0,\sfvar[Q(\bX_{i},d)|S_{i}=0]);\\
				\sqrt{n_{0}}(\widehat{\theta}_{1,\rm eff}^{(1)} - \theta_{1}) \underset{n_{0}\to\infty}{\stackrel{\cD}{\longrightarrow}} \cN(0,\sfvar[C(\bX_{i})d(\bX_{i})|S_{i}=0]); \quad  \sqrt{n_{0}}(\widehat{\theta}_{1,\rm eff}^{(2)} - \theta_{1}) \underset{n_{0}\to\infty}{\stackrel{\cD}{\longrightarrow}} \cN(0,\sfvar[C(\bX_{i})d(\bX_{i})|S_{i}=0]),
			\end{array}
		\end{align*}
		which attain the corresponding semiparametric $ \sqrt{n_{0}} $-asymptotic variance lower bounds for $ \theta $ and $ \theta_{1} $ respectively.
	\end{coro}
	
	\noindent Notice that the $ \sqrt{n_{0}} $-asymptotic variances from Corollary \ref{coro:eff} are the same as $ {1 \over n_{0}}\sum_{i:S_{i}=0}Q(\bX_{i},d) $ and $ {1 \over n_{0}}\sum_{i:S_{i}=0}C(\bX_{i})d(\bX_{i}) $ respectively, so that all efficiency augmentation terms in (\ref{eq:oracle}) are negligible with respect to the $ \sqrt{n_{0}} $-order. Moreover, any estimate $ \widehat{Q}(\bx,a) $ of $ Q(\bx,a) $ that satisfies $ \| \widehat{Q}(\cdot,a) - Q(\cdot,a) \|_{L^{2}(\bbP)} = \smallcO_{\bbP}(n_{0}^{-1/2}) $ for $ a \in \{1,-1\} $ will not affect these $ \sqrt{n_{0}} $-asymptotic results. Since we have assumed that $ n_{0}/n_{1} \to 0 $, it can be relatively easier than Theorem \ref{thm:equiv} to obtain an estimate $ \widehat{Q} $ of order $ \smallcO_{\bbP}(n_{0}^{-1/2}) $ based on the training data. 
	
	\subsection{Estimating Weights}
	In Section \ref{sec:inference}, we obtain the efficient estimates for $ \theta $ and $ \theta_{1} $ on two types of pooled data $ \cO^{(1)} $ and $ \cO^{(2)} $ as in (\ref{eq:oracle}). We also establish the asymptotic properties in Theorems \ref{thm:eff} and \ref{thm:equiv} for the oracle and cross-fitting efficient estimates respectively. In this section, we discuss different methods from literature for estimating training sample weights $ \{w(\bX_{i}): S_{i} = 1\} $ or the nuisance function $ \bx \mapsto w(\bx) $. In particular, \textit{Augmented Inverse Probability of Sampling Weight (AIPSW)} and density ratio estimation are related to the discussions in \citet{li2020discussion} and \citet{liang2020discussion} respectively.
	%and \textit{Augmented Calibration Weight (ACW)} can have an interesting connection to \citet{mo2020learning} when estimating weights.
	
	\paragraph{Augmented Inverse Probability of Sampling Weight (AIPSW)}
	Let $ \widehat{\pi}_{S}(1|\bx) $ be an estimate of $ \pi_{S}(1|\bx) $. Define the estimated weights at training data points $ \{ i:S_{i} = 1 \} $ as
	\[ \widehat{w}(\bX_{i}) := {n_{1}\widehat{\pi}_{S}(0|\bX_{i}) \over n_{0}\widehat{\pi}_{S}(1|\bX_{i})}; \quad \forall i: S_{i}=1. \]
	Then the \textit{AIPSW estimates} \citep{stuart2011use,buchanan2018generalizing} are defined as in (\ref{eq:oracle}) with the AIPSWs $ \{\widehat{w}(\bX_{i}):S_{i} = 1\} $ that are obtained from the estimated conditional-on-covariate sampling probability function $ \widehat{\pi}_{S}(s|\bx) $. We further denote the AIPSW estimates as $ \widehat{\theta}_{\rm AIPSW}^{(1)} $, $ \widehat{\theta}_{\rm AIPSW}^{(2)} $, $ \widehat{\theta}_{1,\rm AIPSW}^{(1)} $ and $ \widehat{\theta}_{1,\rm AIPSW}^{(2)} $ respectively. Notice that the AIPSW estimates have one-to-one correspondence to the estimators proposed in \citet[Section 3.2]{li2020discussion}: $ \widehat{\theta}_{\rm AIPSW}^{(1)} = \widehat{V}_{\rm eff}^{*}(d) $, $ \widehat{\theta}_{\rm AIPSW}^{(2)} = \widehat{V}_{\rm onlyX}^{*}(d) $, $ \widehat{\theta}_{1,\rm AIPSW}^{(1)} = \widehat{R}_{\rm eff}^{*}(d) $, and $ \widehat{\theta}_{1,\rm AIPSW}^{(2)} = \widehat{R}_{\rm onlyX}^{*}(d) $. %This is not surprising since these estimates are obtained from the corresponding EIFs. 
	
	In practice, $ \widehat{\pi}_{S}(s|\bx) $ can be estimated by logistic regression of $ S_{i} $ on $ \bX_{i} $. Consider the simulation setup in \citet{li2020discussion}: $ \bX_{i}|(S_{i} = 1) \sim \cN_{p}(\bzero,\Ib_{p}) $ and $ \bX_{i}|(S_{i} = 0) \sim \cN_{p}(\bmu,\Ib_{p}) $ for some $ \bmu \in \bbR^{p} $. For $ s \in \{1,0\} $, denote $ f(\bx|s) $ as the density of $ \bX_{i}|(S_{i}=s) $. The theoretical weight function in this case is $ w(\bx) = {f_{X}(\bx|0) \over f_{X}(\bx|1)} = \exp(\|\bmu\|_{2}^{2}/2-\bmu^{\intercal}\bx) $. The corresponding log-odds of $ S_{i}|(\bX_{i} = \bx) $ is: $ \log\left( {\pi_{S}(1|\bx)\over \pi_{S}(0|\bx)} \right) = \log\left( {\rho_{S} \over 1 - \rho_{S}} \right) -\log[w(\bx)] = \log\left( {\rho_{S}\over 1-\rho_{S}} \right) - \|\bmu\|_{2}^{2}/2 + \bmu^{\intercal}\bx $. In this case, logistic regression can correctly specify $ S_{i}|\bX_{i} $. Therefore, the corresponding AIPSW estimates can enjoy semiparametric efficiency, which was illustrated in the numerical studies of \citet[Section 3.3]{li2020discussion}.
	
	\paragraph{Density Ratio Estimation}
	Instead of estimating $ \pi_{S}(s|\bx) $ first to obtain $ w(\bx)  = {\rho_{S} \over 1 - \rho_{S}}{\pi_{S}(0|\bx) \over \pi_{S}(1|\bx)} $, we can directly consider $ w(\bx) $ as the covariate density ratio function $ {\bbP(\bX_{i} = \bx|S_{i}=0) \over \bbP(\bX_{i} = \bx|S_{i}=1)} $. \citet{uehara2020off} proposed to estimate the covariate density ratio for $ \{ \bX_{i}:S_{i}=1 \} $ using the \textit{Kernel-based unconstrained Least-Squares Importance Fitting (KuLSIF)} \citep{kanamori2012statistical}, which was also discussed in \citet{liang2020discussion}. Specifically, let $ \cG $ be a generic function class, and consider the following least-squared problem:
	\[ \min_{g \in \cG}{1 \over 2}\bbE\Big\{ [g(\bX_{i}) - w(\bX_{i})]^{2}\Big|S_{i} = 1 \Big\} = \min_{g \in \cG}\left\{ {1 \over 2}\bbE[g(\bX_{i})^{2}|S_{i}=1] - \bbE[g(\bX_{i})|S_{i}=0] \right\} + {1 \over 2}\bbE[w(\bX_{i})^{2}|S_{i}=1]. \]
	%The optimization for $ g \in \cG $ is ``unconstrained'' since the theoretical covariate weight function satisfies $ w(\bX_{i}) \ge 0 $ and $ \bbE[w(\bX_{i})|S_{i} = 1] = 1 $, while the above least-squared problem does not incorporate such constraints. 
	The empirical version of the above least-squared problem is as follows:
	\begin{align}
		\min_{g \in \cG}\left\{ {1 \over 2 n_{1}}\sum_{i:S_{i}=1}g(\bX_{i})^{2} - {1 \over n_{0}}\sum_{i:S_{i}=0}g(\bX_{i}) + {\lambda \over 2}\| g \|_{\cG}^{2} \right\},\label{eq:KuLSIF}
	\end{align}
	where $ (\lambda/2)\|g\|_{\cG}^{2} $ is a functional penalty on $ \cG $. Let $ K $ be a positive semi-definite kernel function on $ \cX $ and $ \cG = \cH_{K} $ be the corresponding reproducing kernel Hilbert space (RKHS). Define $ \Kb_{11} := [K(\bX_{i},\bX_{j}): S_{i} = S_{j} = 1] $ and $ \Kb_{01} := [K(\bX_{i},\bX_{j}): S_{i} = 0, S_{j} = 1] $. The solution to (\ref{eq:KuLSIF}) can be represented as $ g_{\balpha}(\cdot) = \sum_{i:S_{i}=1}\alpha_{i}K(\bX_{i},\cdot) + [1/(\lambda n_{0})]\sum_{i:S_{i}=0}K(\bX_{i},\cdot) $ \citep[Theorem 1]{kanamori2012statistical}, and the dual optimization problem for $ \balpha = (\alpha_{i}:S_{i}=1)^{\intercal} \in \bbR^{n_{1}} $ is:
	\[ \min_{\balpha \in \bbR^{n_{1}}}\left\{ {1 \over 2}\balpha^{\intercal}\left( {1 \over n_{1}}\Kb_{11} + \lambda\Ib_{n_{1}} \right)\balpha - {1 \over \lambda n_{0}n_{1}}\bone_{n_{0}}^{\intercal}\Kb_{01}\balpha \right\}. \]
	Let $ \widehat{w}(\bx) $ be the KuLSIF estimate of the covariate weight function. Under certain conditions, \citet[Theorem 2]{kanamori2012statistical} showed that $ \|\widehat{w} - w\|_{L^{2}(\bbP)} = \cO_{\bbP}\left( (n_{1}\wedge n_{0})^{-1/(2+\gamma)} \right) $ for some $ \gamma \in (0,2) $.
	
	Based on the KuLSIF weights, \citet{liang2020discussion} considered $ \widehat{\theta}_{1} := (1/n_{1})\sum_{i:S_{i}=1}\widehat{w}(\bX_{i})\widehat{C}(\bX_{i})d(\bX_{i}) $. Under Assumptions \ref{asm:consistency}-\ref{asm:intble}, it requires that $ \|\widehat{w}-w\|_{L^{2}(\bbP)} = \smallcO_{\bbP}(n^{-1/2}) $ and $ \|\widehat{C} - C\|_{L^{2}(\bbP)} = \smallcO_{\bbP}(n^{-1/2}) $ to establish the $ \sqrt{n} $-consistency of $ \widehat{\theta}_{1} $. However, the KuLSIF estimate of $ w(\bx) $ cannot satisfy this rate condition. Even when $ \|\widehat{w}-w\|_{L^{2}(\bbP)} = \smallcO_{\bbP}(n^{-1/2}) $ and $ \|\widehat{C} - C\|_{L^{2}(\bbP)} = \smallcO_{\bbP}(n^{-1/2}) $ so that $ \widehat{\theta}_{1} $ is $ \sqrt{n} $-consistent, the asymptotic variance of $ \widehat{\theta}_{1} $ is generally greater than the semi-parametric efficiency bound in Theorem \ref{thm:eff}. In order to achieve the efficiency bound, the KuLSIF density ratio estimate should be combined with a semiparametric efficient estimate and the cross-fitting strategy. 
	
	\paragraph{Augmented Calibration Weight (ACW)} Let $ \cG $ be a function class on $ \cX $. Lemma \ref{lem:weight} can also motivate the following covariate balancing conditions among covariate functions in $ \cG $ for the training sample weights $ \{W_{i}:S_{i}=1\} $ \citep{imai2014covariate}:
	\begin{align}
		\underbrace{\sum_{i:S_{i}=1}W_{i}g(\bX_{i})}_{\text{empirical version of $ \bbE[w(\bX_{i})g(\bX_{i})|S_{i}=1] $}} = \underbrace{{1 \over n_{0}}\sum_{i:S_{i}=0}g(\bX_{i})}_{\text{empirical version of $ \bbE[g(\bX_{i})|S_{i}=0] $}}; \quad \forall g \in \cG.\label{eq:balance}
	\end{align}
	Based on the balancing conditions (\ref{eq:balance}), \citet{hainmueller2012entropy} proposed to solve the entropy balancing problem for calibration weights:
	\begin{align}
		\begin{array}{cll}
			\displaystyle\min_{W_{i}:S_{i}=1} & \displaystyle\sum_{i:S_{i}=1}W_{i}\log W_{i},\\
			\text{subject to} & W_{i} \ge 0; & \text{for all $ i $ with $ S_{i} = 1 $};\\
			& \displaystyle \sum_{i:S_{i}=1}W_{i} = 1;\\
			& \displaystyle \sum_{i:S_{i}=1}W_{i}g(\bX_{i}) = {1 \over n_{0}}\sum_{i:S_{i}=0}g(\bX_{i}); & \forall g \in \cG.
		\end{array}\label{eq:entropy}
	\end{align}
	Here, the objective function is the empirical Kullback–Leibler divergence of $ \bbP(\cdot|S_{i}=0) $ with respect to $ \bbP(\cdot|S_{i}=1) $, and $ W_{i} $ is the covariate density ratio at $ \bX_{i} $. Therefore, the optimization problem (\ref{eq:entropy}) seeks for the balancing weights $ \{ \widehat{W}_{i}:S_{i}=1 \} $ that satisfies (\ref{eq:balance}) and minimizes the discrepancy of the testing covariate distribution from training. If $ \cG = \{ g_{j} \}_{j=1}^{m} $ and denote $ \bg = (g_{1},g_{2},\cdots,g_{m})^{\intercal} $ as an $ m $-dimensional vector of instrumental functions, then the calibration weights can be solved by the following dual problem:
	\[ \widehat{W}_{i} = {\exp[\widehat{\blambda}^{\intercal}\bg(\bX_{i})] \over \sum_{j:S_{j}=1}\exp[\widehat{\blambda}^{\intercal}\bg(\bX_{j})]}; \quad \text{for all $ i $ with $ S_{i} = 1 $}, \]
	where $ \widehat{\blambda} \in \bbR^{m} $ is solved by the balancing equations (\ref{eq:balance}). The dual solution can be further extended to the case when $ \cG $ is an RKHS \citep{zhao2019covariate}. The calibration weights from (\ref{eq:entropy}) can also correspond to a parametric model for $ w(\bX_{i}) $. Specifically, if $ w(\bx;\bmeta) = \exp[\bmeta^{\intercal}\bg(\bx)] $, then $ \widehat{\blambda} \stackrel{\bbP}{\to} \bmeta $, and $ \widehat{W}_{i} = {w(\bX_{i}) \over n_{1}} + \smallcO_{\bbP}(n^{-1}) $ \citep[Theorem 1]{dong2020integrative}. The following two cases can imply such a parametric model:
	\begin{enumerate}[label=(\Roman*)]
		\item Logistic regression of $ S_{i} $ on $ \bg(\bX_{i}) $ can imply such a parametric model for $ w(\bX_{i}) $;
		
		\item If $ \bg(\bX_{i}) $ is a sufficient statistic for $ \bX_{i} $ and $ \bg(\bX_{i})|(S_{i} = s) \sim \cN_{m}(\bmu_{s},\Sigma) $ for $ s \in \{1,0\} $, then $ w(\bx) = \exp\left\{ (\bmu_{1} - \bmu_{0})^{\intercal}\left( \bg(\bx) - {\bmu_{1} + \bmu_{0} \over 2} \right) \right\} $. Because of the additional term $ {\bmu_{1} + \bmu_{0} \over 2} $ in $ w(\bx) $, a constant function is required in the function class $ \cG $ for balancing conditions.
	\end{enumerate}
	Finally, the \textit{ACW estimates} are defined as the efficient estimates (\ref{eq:oracle}) with $ \{ w(\bX_{i}):S_{i}=1 \} $ as $ \{ n_{1}\widehat{W}_{i}: S_{i}=1 \} $ from (\ref{eq:entropy}) \citep{dong2020integrative}. One advantage of such estimates is the implicit nonparametric function class specification for $ w(\bx) $.

	\subsection{Challenges for Efficient Policy Evaluation}
	To summarize, we have studied the policy evaluation problem when a set of calibrating data from the testing distribution can be used for training. We establish the efficient policy evaluation results based on the EIFs and discussed the properties of efficient estimates that can be related to existing literature and the discussions in \citet{li2020discussion,liang2020discussion}. However, when the calibrating sample size $ n_{0} $ is small, our discussion suggests the following two main challenges:
	
	\begin{enumerate}[label=(\Roman*)]
		\item Efficient policy evaluation requires a sufficiently large calibrating sample size for useful efficiency gain. Assumption \ref{asm:pos_s} can imply that the asymptotic sampling rates $ n_{1}/n \to \bbP(S_{i} = 1) $ and $ n_{0}/n \to \bbP(S_{i} = 0) $ are both at least $ \delta $, which requires that both the training and calibrating sample sizes $ n_{1} $ and $ n_{0} $ grow linearly in $ n $. 
		When $ n_{0} = \smallcO(n_{1}) $, Corollary \ref{coro:eff} suggests that the $ \sqrt{n_{0}} $-asymptotic efficient estimates in (\ref{eq:oracle}) are equivalent to averaging the nonparametric estimates of $ Q(\bX_{i},d) $ or $ C(\bX_{i})d(\bX_{i}) $ over the calibrating data. The complicated forms in (\ref{eq:oracle}) may not be helpful for efficiency improvement in this case. More importantly, the resulting estimates can be unstable due to the limited calibrating sample size $ n_{0} $;
		
		\item Nuisance function estimates can be hard to obtain in efficient policy evaluation. Specifically, the optimality of efficient estimates requires that the plug-in nuisance function estimates $ \widehat{w}(\bx) $, $ \widehat{\pi}_{A}(a|\bx,s) $ and $ \widehat{Q}(\bx,a) $ are $ \sqrt{n} $-negligible as in Theorem \ref{thm:equiv}. However, the challenge for nuisance function estimation mainly appears in the covariate weight function $ \widehat{w}(\bx) $, since its rate of convergence is determined by $ \min\{n_{1},n_{0}\} $. The same difficulty can appear when estimating the conditional-on-covariate selection probability function $ \pi_{S}(s|\bx) $ in the AIPSW estimates, where a correctly specified parametric estimate $ \widehat{\pi}_{S}(s|\bx) $ is in the $ (n_{1}\wedge n_{0})^{1/2} $-order. Thus, it can be difficult to estimate the nuisance function when $ n_{0} $ is small.
	\end{enumerate}

	Given the challenges of efficient policy evaluation with a limited calibrating sample size $ n_{0} $, DRITR can be less dependent on $ n_{0} $. First of all, DRITR utilizes the training data only to obtain the set of candidates $ \{ \widehat{d}_{c} \}_{c \in \cC} $, where $ \cC $ is a set of candidate DR-constants, and each $ \widehat{d}_{c} $ enjoys certain robust performance guarantee. Then calibrating sample is used to choose the final DRITR. In this way, DRITR is less affected by the size of $ n_{0} $ compared to the combined strategy in Section \ref{sec:evaluation}.
	Secondly, the value function estimates used in evaluating candidate DRITRs on the calibrating dataset can still enjoy certain properties with a small $ n_{0} $. In \citet[Section 2.4]{mo2020learning}, two calibration procedures were proposed, one using the calibrating covariates only, and another one using the calibrating covariates, treatments and outcomes. In the calibration procedure based on calibrating covariates only, the testing value function estimate for a given ITR $ d $ is $ {1 \over n_{0}}\sum_{i:S_{i}=0}\widehat{C}(\bX_{i})d(\bX_{i}) $, which is semiparametric $ \sqrt{n_{0}} $-asymptotic efficient due to Corollary \ref{coro:eff}. For another calibration procedure based on calibrating covariates, treatments and outcomes, the testing value function estimate is $ {1 \over n_{0}}\sum_{i:S_{i}=0}{\bbone[d(\bX_{i})=A_{i}] \over \pi_{A}(A_{i}|\bX_{i},1)}Y_{i} $. Although such an estimate may not achieve the $ \sqrt{n_{0}} $-variance lower bound, it can be robust if Assumption \ref{asm:mean_s} is violated. Specifically, if Assumption \ref{asm:mean_s} is violated, the estimates (\ref{eq:oracle}) may not be consistent, while $ {1 \over n_{0}}\sum_{i:S_{i}=0}{\bbone[d(\bX_{i})=A_{i}] \over \pi_{A}(A_{i}|\bX_{i},1)}Y_{i} $ remains $ \sqrt{n_{0}} $-consistent. Thirdly, we would like to point out that DRITR avoids estimating the covariate weight function $ w(\bx) $. Therefore, it can bypass the challenge of nuisance function estimation given the limited calibrating data.
	
	\section{Applicability of DRITR} \label{sec:app}
	
	In Section \ref{sec:compare}, we distinguish DRITR from retargeted policy learning as it focuses on covariate changes. In Section \ref{sec:evaluation}, we consider the problem of covariate changes with calibrating data from a specific testing distribution being available at the training stage. In particular, we discuss the general challenges for efficient policy evaluation when available information from testing is limited, and how DRITR can avoid such challenges. As is mentioned at the beginning of Section \ref{sec:evaluation}, DRITR focused on general covariate changes instead of a specific testing distribution. In this section, we discuss the general applicability of DRITR.
	
	We first emphasize that DRITR aims for protecting scientific discoveries from the general agnostic covariate changes. This explains why, in response to \citet{li2020discussion}, we proposed to work with the least favorable case among some possible covariate changes. In fact, the concern on potential training-testing distributional changes can be important in modern prediction methodology. \citet[Section 6]{efron2020prediction} discussed their analysis on the prostate cancer microarray study. If they randomly split data into training and testing, then the testing error of a random forest classifier can be as low as 2\%. However, if they selected patients with the lowest ID numbers into the training dataset, with the remaining for testing purpose, then the testing error would be as high as 24\%. We also performed similar analysis on the ACTG175 study. In particular, we found that when testing on the female population, several other existing methods can have poorer performance than DRITR. Such a violation of the identical training and testing distributions can undermine an existing scientific finding, and researchers may question the faithfulness of such a finding when generalizing it to a much broader scope. On the contrary, a scientific finding robust to all such violations can typically be closer to universal, eternal truths and become long-lasting \citep{efron2020prediction}. The same scientific principle has also been advocated in \citet{yu2020veridical,buhlmann2020invariance}, both of which established nice connections of such a principle with adversarial perturbation and distributional robustness.
	
	DRITR can correspond to a more ``forgiving'' but useful approach than precise estimation. On one hand, we agree with \citet{dukes2020discussion} that making correct ``causal predictions'', that is, estimating the CTE function correctly, can be the most robust way of protecting from covariate changes. In fact, we highlight in our Section \ref{sec:compare} that general applicability of DRITR relies on the assumption $ \{ \bx \mapsto \sfsign[C(\bx)] \} \nsubseteq \cD $. One example is when the true CTE function $ C(\bx) $ is too complicated to be estimable, the ITR class $ \cD $ can be misspecified. Another example is that the CTE function $ C(\bx) $ takes a complicated functional form, and $ \cD $ is intended for a more parsimonious class of decision rules in practice. In either of these two cases, DRITR can be a useful methodology with tolerance on incorrect ``causal predictions''. On the other hand, given our combined data analysis in Section \ref{sec:evaluation}, efficient inference of parameters of interests can have more restrictive requirements on data availability and involve more assumptions. In contrast, the requirements for accurate predictions, \textit{e.g.}, predicting which treatment to assign in our context, can typically be less stringent than drawing efficient inference of parameter estimates, as was discussed in \citet[Criteria 6]{efron2020prediction}. These can distinguish the prediction-driven focus and usefulness of DRITR. While an inference-based criterion can only be applicable if all required assumptions hold, a prediction-based criterion particularly focuses on some measurements of testing performance and can be less restrictive. Therefore, even though DRITR can be conservative by performing worst-case policy optimization,
%	compared to any specific testing distribution and any policy evaluation approaches that make use of training and calibrating data efficiently, 
	it can enjoy less restrictions and more general applicability.
	
	The last point we would like to point out is that the training of candidate DRITRs can be performed before using calibrating data. This can provide more privacy protection. Specifically, DRITR can utilize the training data to obtain a class of candidate ITR estimates $ \{ \widehat{d}_{c} \}_{c \in \cC} $, where $ \cC $ is the set of candidate DR-constants. When estimating the optimal DRITR on a specific testing distribution, we only use the testing information to choose the best ITR from $ \{ \widehat{d}_{c} \}_{c \in \cC} $ without requesting for the complete training data. In contrast, the combined analysis in Section \ref{sec:evaluation} requires at least either $ \{\bX_{i},\widehat{Q}(\bX_{i},\pm 1):S_{i}=1\} $ or $ \{\bX_{i},\widehat{C}(\bX_{i}):S_{i}=1\} $ from the training data. In this case, treatment effect information at the individual level would be exposed to the testing agents. Therefore, the individualized treatment effect information obtained from training can be kept privately when applying DRITR, but cannot when using methods based on combined data in Section \ref{sec:evaluation}.
	
%	\section*{Appendix}
%	\addcontentsline{toc}{section}{Appendix}
%	\renewcommand{\thesection}{A}
%	\begin{proof}[Proof of Lemma \ref{lem:weight}]
%		\begin{align*}
%			\bbE[g(\bX)|S=0] &= \bbE\left\{ {\bbone(S=0) \over \bbP(S=0)}\times g(\bX) \right\}\\
%			&= \bbE\left\{ {\bbP(S=0|\bX) \over \bbP(S=0)} \times {\bbone(S=0) \over \bbP(S=0|\bX)}\times g(\bX) \right\}\\
%			&= \bbE\left\{ {\bbP(S=0|\bX) \over \bbP(S=0)} \times g(\bX) \right\}\\
%			&= \bbE\left\{ {\bbP(S=0|\bX) \over \bbP(S=0)} \times {\bbone(S=1) \over \bbP(S=1|\bX)}\times g(\bX) \right\}\\
%			&= \bbE\left\{ {\bbP(S=1)\bbP(S=0|\bX) \over \bbP(S=0)\bbP(S=1|\bX)} \times {\bbone(S=1) \over \bbP(S=1)}\times g(\bX) \right\}\\
%			&= \bbE[w(\bX)g(\bX)|S=1].
%		\end{align*}
%		Finally, let $ f_{\bX|S}(\bx|s) $ be the probability density of $ \bX|(S=s) $ for $ s \in \{1,0\} $. Then by Bayes rule, we have $ w(\bX) = f_{\bX|S}(\bX|0)/f_{\bX|S}(\bX|1) $. Therefore, we have
%		\[ \bbE[w(\bX)|S=1] = \int_{\cX}{f_{\bX|S}(\bx|0) \over f_{\bX|S}(\bx|1)}\times f_{\bX|S}(\bx|1)\rd\bx = \int_{\cX}f_{\bX|S}(\bx|0)\rd \bx = 1. \]
%	\end{proof}

	\bibliography{bibfile.bib}

\begin{thebibliography}{29}
\newcommand{\enquote}[1]{``#1''}
\expandafter\ifx\csname natexlab\endcsname\relax\def\natexlab#1{#1}\fi

\bibitem[{Athey and Wager(2020)}]{athey2020policy}
Athey, S. and Wager, S. (2020), \enquote{Policy learning with observational
  data,} \textit{Econometrica}, to appear.

\bibitem[{Buchanan et~al.(2018)Buchanan, Hudgens, Cole, Mollan, Sax, Daar,
  Adimora, Eron, and Mugavero}]{buchanan2018generalizing}
Buchanan, A.~L., Hudgens, M.~G., Cole, S.~R., Mollan, K.~R., Sax, P.~E., Daar,
  E.~S., Adimora, A.~A., Eron, J.~J., and Mugavero, M.~J. (2018),
  \enquote{Generalizing evidence from randomized trials using inverse
  probability of sampling weights,} \textit{Journal of the Royal Statistical
  Society: Series A (Statistics in Society)}, 181, 1193--1209.

\bibitem[{B{\"u}hlmann(2020)}]{buhlmann2020invariance}
B{\"u}hlmann, P. (2020), \enquote{Invariance, causality and robustness,}
  \textit{Statistical Science}, 35, 404--426.

\bibitem[{Chernozhukov et~al.(2018)Chernozhukov, Chetverikov, Demirer, Duflo,
  Hansen, Newey, and Robins}]{chernozhukov2018double}
Chernozhukov, V., Chetverikov, D., Demirer, M., Duflo, E., Hansen, C., Newey,
  W., and Robins, J. (2018), \enquote{Double/debiased machine learning for
  treatment and structural parameters,} \textit{The Econometrics Journal}, 21,
  C1--C68.

\bibitem[{Crump et~al.(2006)Crump, Hotz, Imbens, and Mitnik}]{crump2006moving}
Crump, R.~K., Hotz, V.~J., Imbens, G.~W., and Mitnik, O.~A. (2006),
  \enquote{Moving the goalposts: Addressing limited overlap in the estimation
  of average treatment effects by changing the estimand,} Tech. rep., National
  Bureau of Economic Research.

\bibitem[{Crump et~al.(2009)Crump, Hotz, Imbens, and Mitnik}]{crump2009dealing}
--- (2009), \enquote{Dealing with limited overlap in estimation of average
  treatment effects,} \textit{Biometrika}, 96, 187--199.

\bibitem[{Dahabreh et~al.(2019)Dahabreh, Robertson, Petito, Hern{\'a}n, and
  Steingrimsson}]{dahabreh2019efficient}
Dahabreh, I.~J., Robertson, S.~E., Petito, L.~C., Hern{\'a}n, M.~A., and
  Steingrimsson, J.~A. (2019), \enquote{Efficient and robust methods for
  causally interpretable meta-analysis: transporting inferences from multiple
  randomized trials to a target population,} \textit{arXiv preprint
  arXiv:1908.09230}.

\bibitem[{Dong et~al.(2020)Dong, Yang, Wang, Zeng, and
  Cai}]{dong2020integrative}
Dong, L., Yang, S., Wang, X., Zeng, D., and Cai, J. (2020),
  \enquote{Integrative analysis of randomized clinical trials with real world
  evidence studies,} \textit{arXiv preprint arXiv:2003.01242}.

\bibitem[{Dukes and Vansteelandt(2020)}]{dukes2020discussion}
Dukes, O. and Vansteelandt, S. (2020), \textit{Discussion of Kallus and Mo, Qi,
  and Liu: New Objectives for Policy Learning}.

\bibitem[{Efron(2020)}]{efron2020prediction}
Efron, B. (2020), \enquote{Prediction, Estimation, and Attribution,}
  \textit{Journal of the American Statistical Association}, 115, 636--655.

\bibitem[{Hainmueller(2012)}]{hainmueller2012entropy}
Hainmueller, J. (2012), \enquote{Entropy balancing for causal effects: A
  multivariate reweighting method to produce balanced samples in observational
  studies,} \textit{Political analysis}, 20, 25--46.

\bibitem[{Imai and Ratkovic(2014)}]{imai2014covariate}
Imai, K. and Ratkovic, M. (2014), \enquote{Covariate balancing propensity
  score,} \textit{Journal of the Royal Statistical Society: Series B:
  Statistical Methodology}, 76, 243--263.

\bibitem[{Kallus(2020)}]{kallus2020more}
Kallus, N. (2020), \enquote{More Efficient Policy Learning via Optimal
  Retargeting,} \textit{Journal of the American Statistical Association}, to
  appear.

\bibitem[{Kanamori et~al.(2012)Kanamori, Suzuki, and
  Sugiyama}]{kanamori2012statistical}
Kanamori, T., Suzuki, T., and Sugiyama, M. (2012), \enquote{Statistical
  analysis of kernel-based least-squares density-ratio estimation,}
  \textit{Machine Learning}, 86, 335--367.

\bibitem[{Li et~al.(2020)Li, Li, and Luedtke}]{li2020discussion}
Li, S., Li, X., and Luedtke, A. (2020), \enquote{Discussion of Kallus (2020)
  and Mo, Qi, and Liu (2020): New Objectives for Policy Learning,}
  \textit{arXiv preprint arXiv:2010.04805}.

\bibitem[{Liang and Zhao(2020)}]{liang2020discussion}
Liang, M. and Zhao, Y. (2020), \textit{Discussion of Kallus (2020) and Mo et al
  (2020)}.

\bibitem[{Mo et~al.(2020)Mo, Qi, and Liu}]{mo2020learning}
Mo, W., Qi, Z., and Liu, Y. (2020), \enquote{Learning Optimal Distributionally
  Robust Individualized Treatment Rules,} \textit{Journal of the American
  Statistical Association}, to appear.

\bibitem[{Molina et~al.(2017)Molina, Rotnitzky, Sued, and
  Robins}]{molina2017multiple}
Molina, J., Rotnitzky, A., Sued, M., and Robins, J. (2017), \enquote{Multiple
  robustness in factorized likelihood models,} \textit{Biometrika}, 104,
  561--581.

\bibitem[{Pearl and Bareinboim(2014)}]{pearl2014external}
Pearl, J. and Bareinboim, E. (2014), \enquote{External validity: From
  do-calculus to transportability across populations,} \textit{Statistical
  Science}, 29, 579--595.

\bibitem[{Qi et~al.(2019)Qi, Cui, Liu, and Pang}]{qi2019estimation}
Qi, Z., Cui, Y., Liu, Y., and Pang, J.-S. (2019), \enquote{Estimation of
  Individualized Decision Rules Based on an Optimized Covariate-Dependent
  Equivalent of Random Outcomes,} \textit{SIAM Journal on Optimization}, 29,
  2337--2362.

\bibitem[{Robins et~al.(2000)Robins, Rotnitzky, and van~der
  Laan}]{robins2000profile}
Robins, J.~M., Rotnitzky, A., and van~der Laan, M. (2000), \enquote{On profile
  likelihood: comment,} \textit{Journal of the American Statistical
  Association}, 95, 477--482.

\bibitem[{Rubin(1974)}]{rubin1974estimating}
Rubin, D.~B. (1974), \enquote{Estimating causal effects of treatments in
  randomized and nonrandomized studies,} \textit{Journal of Educational
  Psychology}, 66, 688--701.

\bibitem[{Rudolph and van~der Laan(2017)}]{rudolph2017robust}
Rudolph, K.~E. and van~der Laan, M.~J. (2017), \enquote{Robust estimation of
  encouragement-design intervention effects transported across sites,}
  \textit{Journal of the Royal Statistical Society. Series B, Statistical
  methodology}, 79, 1509–1525.

\bibitem[{Stuart et~al.(2011)Stuart, Cole, Bradshaw, and Leaf}]{stuart2011use}
Stuart, E.~A., Cole, S.~R., Bradshaw, C.~P., and Leaf, P.~J. (2011),
  \enquote{The use of propensity scores to assess the generalizability of
  results from randomized trials,} \textit{Journal of the Royal Statistical
  Society: Series A (Statistics in Society)}, 174, 369--386.

\bibitem[{Tsiatis(2007)}]{tsiatis2007semiparametric}
Tsiatis, A. (2007), \textit{Semiparametric theory and missing data}, Springer
  Science \& Business Media.

\bibitem[{Uehara et~al.(2020)Uehara, Kato, and Yasui}]{uehara2020off}
Uehara, M., Kato, M., and Yasui, S. (2020), \enquote{Off-Policy Evaluation and
  Learning for External Validity under a Covariate Shift,} \textit{Advances in
  Neural Information Processing Systems}, to appear.

\bibitem[{Yu and Kumbier(2020)}]{yu2020veridical}
Yu, B. and Kumbier, K. (2020), \enquote{Veridical data science,}
  \textit{Proceedings of the National Academy of Sciences}, 117, 3920--3929.

\bibitem[{Zhao(2019)}]{zhao2019covariate}
Zhao, Q. (2019), \enquote{Covariate balancing propensity score by tailored loss
  functions,} \textit{The Annals of Statistics}, 47, 965--993.

\bibitem[{Zhao et~al.(2019)Zhao, Zeng, Tangen, and
  Leblanc}]{zhao2019robustifying}
Zhao, Y.-Q., Zeng, D., Tangen, C.~M., and Leblanc, M.~L. (2019),
  \enquote{Robustifying trial-derived optimal treatment rules for a target
  population,} \textit{Electronic Journal of Statistics}, 13, 1717--1743.

\end{thebibliography}
	\bibliographystyle{asa}
\end{document}